%% file: main.tex
\documentclass[10pt,twocolumn,letterpaper]{article}

\usepackage{iccv}              


\definecolor{iccvblue}{rgb}{0.21,0.49,0.74}
\usepackage[pagebackref,breaklinks,colorlinks,allcolors=iccvblue]{hyperref}

\def\paperID{2567} 
\def\confName{ICCV}
\def\confYear{2025}

\title{3D Human Mesh Estimation from Single View RGBD}

\author{Ozhan Suat$^1$, Bedirhan Uguz$^1$, Batuhan Karagoz$^1$, Muhammed Can Keles$^1$, Emre Akbas$^{1,2, 3}$ \\
$^1$Department of Computer Engineering and $^2$METU-ROMER Robotics Center\\
Middle East Technical University, Ankara, Turkey\\
$^3$Helmholtz Munich, Germany\\
{\tt\small \{ozhan.suat, bedirhan.uguz, batuhan.karagoz, can.keles, eakbas\}@metu.edu.tr}}

\begin{document}
\maketitle
\input{sec/0_abstract}    
\input{sec/1_intro}
\input{sec/2_related_work}
\input{sec/3_method}
\input{sec/4_implementation_details}
\input{sec/5_experiments}
\input{sec/6_conclusion}

{
    \small
    \bibliographystyle{ieeenat_fullname}
    \bibliography{main}
}

\input{sec/X_supp.tex}

\end{document}

%% file: sec/0_abstract.tex
\begin{abstract}
Despite significant progress in 3D human mesh estimation from RGB images; RGBD cameras, offering additional depth data, remain underutilized. 
In this paper, we present a method for accurate 3D human mesh estimation from a single RGBD view, leveraging the affordability and widespread adoption of RGBD cameras for real-world applications.
A fully supervised approach for this problem, requires a dataset with RGBD image and 3D mesh label pairs. 
However, collecting such a dataset is costly and challenging, hence, existing datasets are small, and limited in pose and shape diversity. 
To overcome this data scarcity, we leverage existing Motion Capture (MoCap) datasets. 
We first obtain complete 3D meshes from the body models found in MoCap datasets, and create partial, single-view versions of them by projection to a virtual camera. 
This simulates the depth data provided by an RGBD camera from a single viewpoint. 
Then, we train a masked autoencoder to complete the partial, single-view mesh. 
During inference, our method, which we name as M$^3$ for ``Masked Mesh Modeling'', matches the depth values coming from the sensor to vertices of a template human mesh, which creates a partial, single-view mesh. 
We effectively recover parts of the 3D human body mesh model that are not visible, resulting in a full body mesh. 
M$^3$ achieves 16.8 mm and 22.0 mm per-vertex-error (PVE) on the SURREAL and CAPE datasets, respectively; outperforming existing methods that use full-body point clouds as input. 
We obtain a competitive 70.9 PVE on the BEHAVE dataset, outperforming a recently published RGB based method by 18.4 mm,
highlighting the usefulness of depth data. 
Code will be released.
\end{abstract}

%% file: sec/1_intro.tex
\section{Introduction}
\label{sec:intro}

Human pose and shape estimation, a fundamental task in computer vision, has many applications across a variety of domains including computer graphics, healthcare, sports, and AR/VR. Significant progress has been made in estimating a 3D human body mesh model from a single RGB image \cite{Bogo:ECCV:2016, kanazawa2018end, lin2021end, ma20233d, luan2021pc, guan2022out, dwivedi_cvpr2024_tokenhmr}. 
These methods use a variety of training cues, including 2D \cite{choi2020pose2mesh, kolotouros2019learning} or 3D keypoints \cite{kanazawa2018end}, body silhouettes \cite{yu2021skeleton2mesh}, and UV maps \cite{guler2019holopose}.
While these methods yield successful results in estimating human body shapes and poses, they do come with certain limitations. 
One notable limitation is the absence of direct depth information in RGB images, which complicates accurate 3D reconstruction, highlighting the main issue of transitioning from 2D to 3D and recovering lost data during projection. 
This limitation becomes especially evident when significant depth variations impact the accurate portrayal of body shapes. 
Additionally, the challenge extends to accurately determining the real-world scale of subjects solely from RGB images, introducing uncertainties in estimating body proportions. 

\begin{figure}
\centering
\includegraphics[width=.43\textwidth]{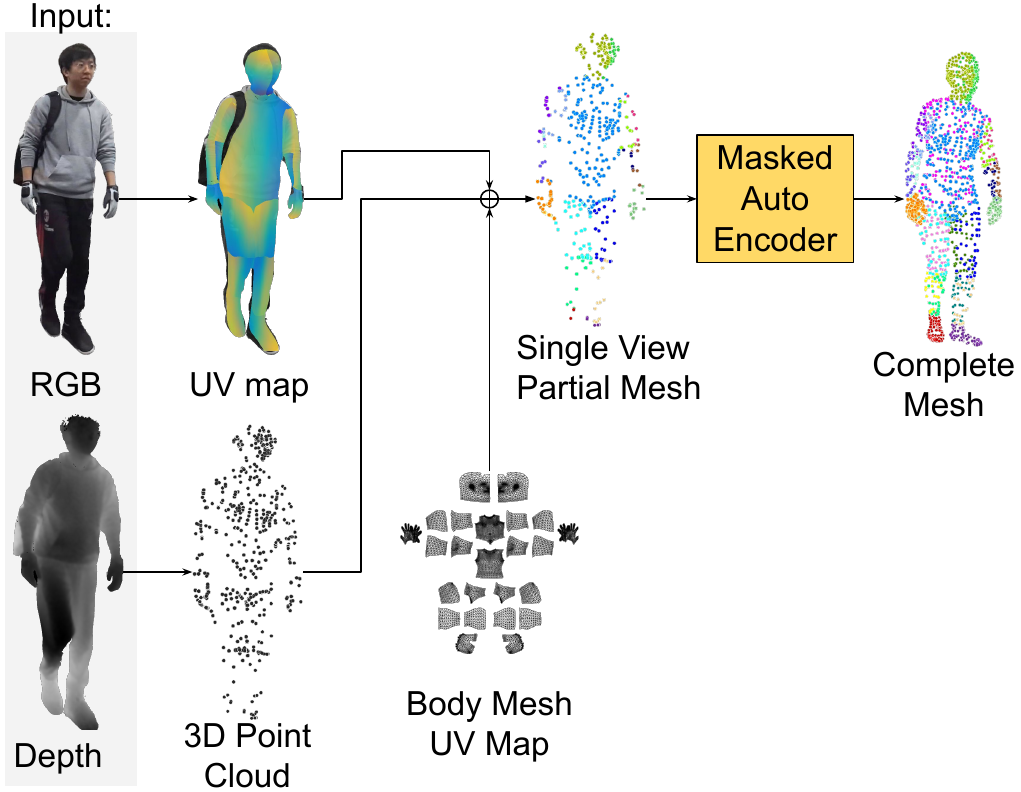}
\caption{Inference pipeline of the proposed method, M$^3$ (short for ``Masked Mesh Modeling''). First, we run an off-the-shelf 2D dense UV estimator on the RGB image. 
Then, we lift the body pixels to 3D points using depth data. 
Next, we match these 3D points to the vertices of a 3D human body mesh model using the UV correspondence between the pixels and the vertices of the template mesh. 
Finally, a masked autoencoder model completes the partial mesh to obtain the full body 3D mesh. Body Mesh UV Map image is taken from DensePose \cite{densepose:Güler_2018_CVPR}.}
\label{fig:inference}
\end{figure}

There are also efforts to estimate 3D body mesh from 3D data. Prominent methods \cite{jiang2019skeleton, bhatnagar2020loopreg} typically rely on 3D scans, introducing challenges related to specialized and potentially costly equipment setups. 
They also involve time-intensive test-time optimization. 
Additionally, there had been efforts to obtain 3D human body pose from single-view RGBD data \cite{shotton2012efficient, ye2011accurate, newcombe2011kinectfusion}. These pre-deep-learning and pre-SMPL \cite{SMPL:2015} methods focus on finding 3D key points from single-view point clouds, and do not estimate a 3D body mesh.

In this paper, we tackle the problem of estimating a 3D human body mesh from a single-view RGBD image (Fig. \ref{fig:inference}). As RGBD cameras have become more affordable and common, this is a realistic, practical problem and a surprisingly underexplored one. Our method is highly applicable to real-world applications using a single, low-cost RGBD camera. 
Our multimodal method solves this problem, resulting in improved accuracy and robustness.
The primary challenges in this problem are twofold: 
\begin{enumerate}
\item[(i)] \emph{Accurate completion of the human mesh from partially visible RGB and depth data.} In addition to the consistent occlusion of the body surface that is not visible by the camera,
body self-occlusions also contribute to the difficulty. 
\item[(ii)] \emph{Lack of training data in the form of RGBD image and 3D mesh label pairs.} Collecting ground-truth 3D human mesh labels is both challenging and costly. Existing datasets are small and limited in environment and body pose, shape diversity. 
\end{enumerate}

To tackle the first challenge, inspired by the idea of Masked Image Modelling \cite{he2022masked}, we use the encoder part of a transformer-based masked autoencoder  to complete a  partial human mesh.
To train this model, we leverage existing large-scale MoCap datasets such as AMASS \cite{AMASS:2019}, which contains a large and diverse set of human body shapes and poses with precise mesh models. However, they lack corresponding RGB inputs, marking the second  challenge. 

We address this problem by extracting a 2D UV map from the input RGB image using an off-the-shelf UV map extractor such as DensePose \cite{densepose:Güler_2018_CVPR}. We obtain a point cloud of the human body with corresponding UV values by combining depth data coming from the ``D" channel of the input and the 2D UV map. Next, we apply a matching algorithm to match the 3D points with the vertices of a template human mesh (e.g. SMPL \cite{SMPL:2015}) to yield the partial body mesh. This partial mesh is fed to the autoencoder (described above) to finally obtain the complete human mesh. 

Our training strategy for the masked autoencoder involves a two-fold process. First, we generate visible parts by projecting ground-truth body meshes (coming from a MoCap dataset) onto a fixed virtual camera, allowing us to extract visible points. Subsequently, a transformer-based model learns the relationships and patterns within the visible data, crucial for the completion process. 

We employ a standard measure for evaluation, namely, ``Per Vertex Error'' (PVE). Our method achieves 16.8mm and 22.0mm PVE on SURREAL and CAPE synthetic datasets, respectively. Real-world scenario testing using the BEHAVE \cite{bhatnagar22behave} dataset resulted in 70.9mm PVE.
For reference, a recently published method (TokenHMR, CVPR2024 \cite{dwivedi_cvpr2024_tokenhmr}) achieves 89.3mm PVE on the same dataset, which clearly shows the benefit of using depth data. 
Notably, our method, which uses only a single RGBD image, also outperforms LoopReg \cite{bhatnagar2020loopreg} that operates on a complete (dense multiview) human body scan, on the BEHAVE dataset benchmark.

\noindent \textbf{Our contributions} are as follows:

\begin{itemize}
\item We demonstrate the potential of using single-view RGBD data for robust 3D human mesh estimation, contributing to more practical and accessible methods in various applications.
\item We introduce a training method that does not require paired RGBD-mesh  data, and  uses a large MoCap dataset instead. In this dataset, we simulate depth measurements by placing a virtual camera around the mesh and creating 3D partial human mesh and full human mesh pairs. This allows us to train a model using simulated data and then use it with 
actual depth (D values) from RGBD data. 
\item We develop a novel model to complete a partial mesh, effectively producing a 3D human mesh model from partially visible data points, using a transformer-based encoder. Our results on synthetic and real datasets show the effectiveness of our method. 
\end{itemize}

%% file: sec/2_related_work.tex
\section{Related Work}
\label{sec:relatedwork}
Our method M$^3$ takes in a single-view RGBD image and produces a full body mesh. To the best of our knowledge, there is only one method (Bashirov et al. \cite{bashirov2021real}) that solves the same exact task, in the literature.  Although they tackle the same task as us, our training settings are completely different. They take a direct supervision approach, where they collect a dataset of 56 people using a calibrated rig of five Kinect sensors. Then, they produce 
“ground truth” using a ``slow per-frame optimization based fitting process". Unlike them, we train M$^3$ exclusively on an existing MoCap dataset without any paired RGB data. The code they provide is a demo that  specifically requires an Azure Kinect. And they only report results on their own dataset. For these reasons, we could not make any comparisons. Below, we review existing work that are related to M$^3$ in several aspects. 

\paragraph{Single-view RGBD input.} Before the rise of deep learning (2012) and the proposal of the parametric body mesh model SMPL \cite{SMPL:2015}, there had been efforts to use single-view RGBD and estimate 3D human pose. 
These methods typically process single-view point clouds to derive 3D human body skeletons. Specifically, Shotton et al. \cite{shotton2012efficient} utilizes body part labels for 3D pose estimation, while Ye et al. \cite{ye2011accurate} leverages pre-computed 3D human poses to enhance 3D human pose estimation from point clouds. On the other hand, other methods such as KinectFusion \cite{newcombe2011kinectfusion} specialize in reconstructing humans in 3D scenes from a single-view point cloud.
Unlike our work, the focus of these methods is not producing a full body mesh. 

\paragraph{2D keypoints input.}  SMPLify \cite{Bogo:ECCV:2016} fits 3D human body mesh to given 2D keypoints. They utilize a loop to fit the body mesh step by step. We leveraged a similar network to create a baseline for our method, which fits body mesh vertices to 3D vertices generated from RGBD. MPT \cite{lin2024mpt} also estimates body meshes from sparse 2D keypoints using a single-pass approach and pre-train on AMASS. Key2Mesh \cite{Uguz_2024_CVPR} follows a similar keypoint-to-mesh paradigm but applies adversarial domain adaptation to bridge the gap between MoCap and visual domains, without requiring 3D labels for the target domain. However, instead of projecting 2D keypoints to a virtual camera, we utilize SMPL vertices to generate partial meshes by masking non-visible vertices from the camera's view, making our method better suited for RGBD input.

\paragraph{UV map based methods.} In UV map-based methods for 3D human body mesh estimation, various strategies leverage RGB images for dense correspondence. Zeng et al. \cite{zeng20203d} and DIMR \cite{wang2023learning} estimate an IUV map and feature vector from an RGB image, transferring features to the IUV space for human mesh estimation. Similarly, Zhang et al. \cite{zhang2020object} estimate a UV map and saliency map from an RGB image, using latent features to estimate 3D locations of human mesh vertices. Densepose2SMPL \cite{gu2022dense} estimates the UV map and then utilizes an optimization loop to fit body mesh to UV estimations.  On the other hand, DaNet \cite{zhang2020densepose2smpl} and DenseRac \cite{xu2019denserac} use the UV map as an intermediate representation and use a single pass approach to estimate human mesh from the UV map. DP3D \cite{shapovalov2021densepose} estimates the UV map and then calculates rigid transformations to pose template mesh using UV values. Dense Depth Alignment (DDA) \cite{karagoz2024dense} also uses the UV map to establish dense correspondences between mesh vertices and pixels, aligning them with reconstructed points from an estimated depth map, and introduces a camera pretraining strategy to improve optimization stability. In our case, UV maps are used to find mesh indices of depth points by leveraging their dense correspondence with the body mesh, enabling us to directly obtain 3D locations of visible mesh vertices.

\paragraph{Part-based methods.} Several approaches \cite{bhatnagar2020loopreg, li2019lbs, wang2021locally, feng2023generalizing, bhatnagar2020ipnet} leverage human body parts to enhance the accuracy of human body mesh estimations. These methods employ feature extraction from point clouds to estimate body parts. Arteq \cite{feng2023generalizing} utilizes SE(3) networks to estimate rigid body part rotations. PTF \cite{wang2021locally} estimates occupancies for individual body parts and fits a human mesh using a loop.  Our method adopts a UV map estimation strategy to establish a mapping between 3D points and the template mesh model. It shares similarities with part-based methods as it provides supervision regarding the spatial locations of 3D points on the human body.

\paragraph{Complete point cloud input.} Methods for estimating 3D human body meshes from point clouds commonly rely on full scans \cite{bhatnagar2020loopreg, jiang2019skeleton, wang2020sequential, li2019lbs, groueix20183d} or multi-view RGBD data \cite{bhatnagar22behave}. LoopReg \cite{bhatnagar2020loopreg} lacks robustness to input noise and necessitates noise-free, complete human body scans during testing. In contrast, our approach operates with partial 3D vertices obtained from single-view RGBD images, enhancing accessibility.

\paragraph{Optimization-based methods.} Some methods \cite{bhatnagar2020loopreg, bhatnagar2020ipnet, wang2021locally, bhatnagar22behave} leverage optimization loops to get the final 3D human body mesh estimation. They require a significant amount of processing time during inference. However, we are applying a single-pass approach to estimate 3D human body mesh from single-view RGBD image. It makes our method more efficient than optimization-based methods.

\paragraph{Sequence input.} Some approaches \cite{jang2023dynamic, wang2020sequential} use a sequence of point clouds as input. Jang et al. \cite{jang2023dynamic} utilize a complete point cloud sequence for pre-training and partial point clouds for training and testing. In contrast, our method operates on a single frame RGBD, eliminating the requirement for temporal information.

%% file: sec/3_method.tex
\section{Method}
\label{sec:method}
We aim to recover the vertices $V \in \mathbb{N}^{N \times 3}$ of a template mesh from RGB image $I\in \mathbb{R}^{W\times H \times 3}$ and depth map $D\in\mathbb{R}^{W\times H}$. First, a dense UV estimator $\Phi$ generates a UV map $U$ of the image $I$. We combine the depth map $D$ and the UV map $U$ to \textit{lift} the body pixels to a point cloud $P$ with UV values attached. Utilizing the correspondence between their UV values, we match the 3D points and the template mesh vertices. Through this matching, we obtain a partial mesh $V^\text{in} \in \mathbb{R}^{N \times 3}$ where only the locations of the visible vertices are valid. We denote the visible vertices obtained this way with a mask $m\in \{0,1\}^N$.
From the partial mesh $V^{\text{in}}$ and visibility mask $m$, a masked autoencoder $\Theta$ recovers the locations of vertices that are not visible and denoises the visible ones. 

While the UV estimator $\Phi$ is readily available, we design a transformer-based masked autoencoder for mesh completion and train it using MoCap data. 

The following sections describe the mesh model we use (Section \ref{subsection:bodymodel}), partial mesh generation (Section \ref{subsection:partialmesh}), our $M^3$ architecture for mesh completion (Section \ref{subsection:mmm}), and training data generation (Section \ref{subsection:trainingdata}). 

\subsection{Body Model}\label{subsection:bodymodel}
We utilize SMPL \cite{SMPL:2015} to define our template mesh. Following previous work \cite{lin2021end}, to reduce the memory and computation requirements, we subsample SMPL template mesh to $N=1723$ vertices. To report our results, we upsample to the original mesh using a learned upsampling layer after our MAE.

\begin{figure}
\centering
\includegraphics[width=\columnwidth]{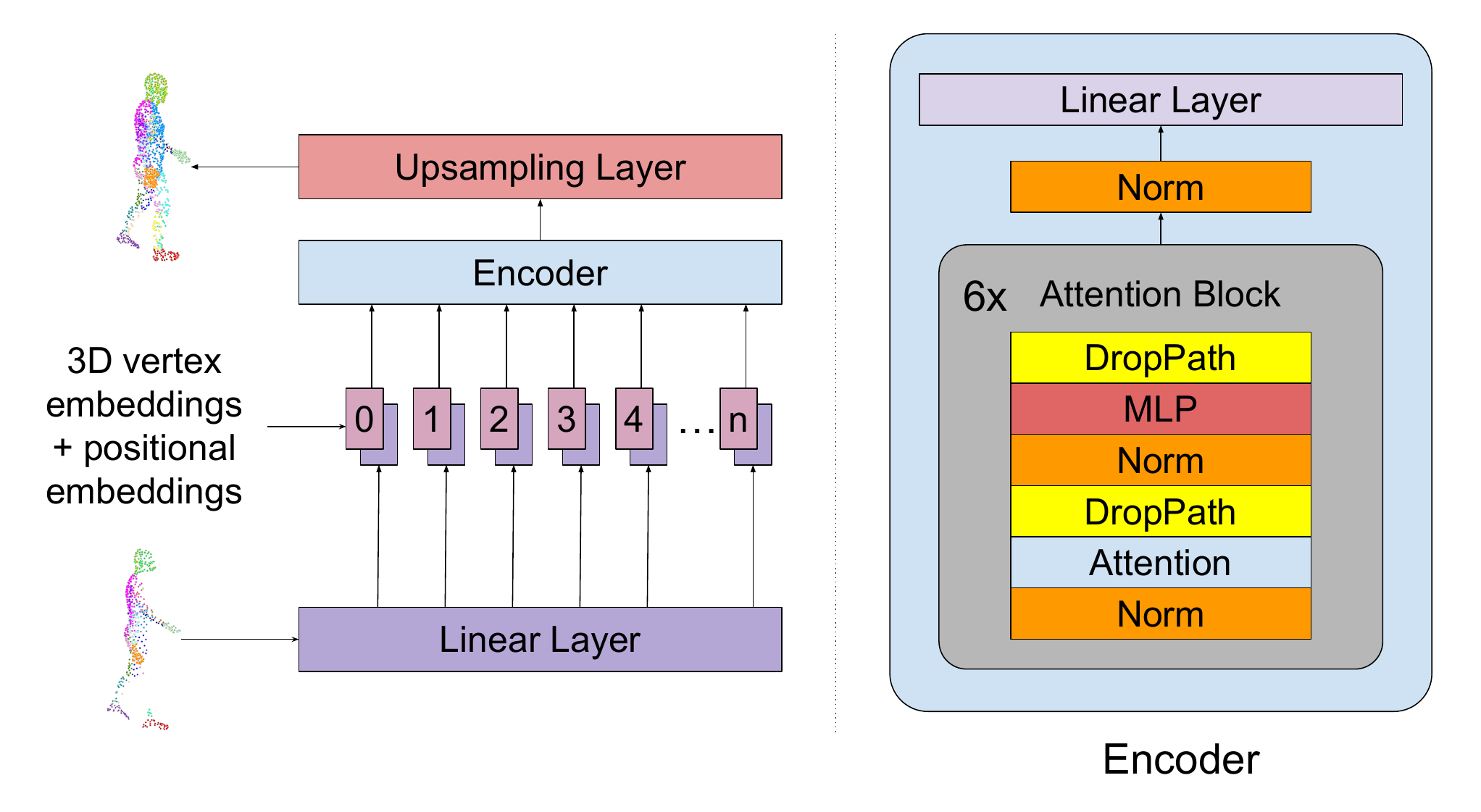}
\caption{Architecture of our transformer-based completion model.}
\label{fig:architecture}
\end{figure}

\subsection{Partial Mesh Generation} \label{subsection:partialmesh}
The UV estimator, $\Phi$, which generates the UV map, $U$, is an instance of DensePose \cite{densepose:Güler_2018_CVPR} with frozen weights. We lift each pixel $p=(x,y)$ belonging to human body to 3D coordinates $\{X(p)\}$ using the intrinsics $K$ and the depth map $D$: 
\begin{equation}
     X(p) = D(p)K^{-1}[x,y,1]^T
 \end{equation}

We leverage the correspondence between the UV values predicted by DensePose and the predefined UV coordinates of SMPL vertices $U^{vert}_i$, where each vertex $i$ on the SMPL model has a unique UV coordinate, to \textit{match} the 3D points $\{X(p)\}$ with the mesh vertices. Using predefined vertex UV values $U^{vert}_i$, we find the point that is closest to the vertex $i$ in terms of UV distance via nearest neighbor search:
\begin{equation}
    V^\text{in}_i = X( \operatorname{argmin}_p||U(p) - U^\mathrm{vert}_i||),
\end{equation}
where $U(p)$ is the estimated UV coordinate of pixel $p$. We mask vertices with the nearest neighbor distance above a threshold to preserve the quality of partial mesh vertices, which means the UV values are considered as \textit{matched} if they are closer than $\epsilon$:
 \begin{equation}
     m_i = \min_{p}||U(p) - U^\mathrm{vert}_i|| < \epsilon
 \end{equation}
 where $V^\text{in}$ is mesh vertices and $m$ is visibility mask. 
  
\subsection{Masked Mesh Modeling} \label{subsection:mmm}

Our masked autoencoder $\Theta$ (Fig. \ref{fig:architecture}) embraces a Transformer-based architecture with mask tokens for non-visible vertices  inspired by ViT-MAE \cite{he2022masked}. The input to $\Theta$ consists of vertices $V^\text{in}$ and their visibility mask $m$. The vertices $V^\text{in}$ are first mapped to 3D location embeddings using a learned linear embedding layer $\mathbf{E}$. A learnable mask token $\textbf{E}_{\text{mask}}$ is used for non-visible vertices. These 3D vertex embeddings after the addition of position embeddings $\mathbf{E}^{\text{pos}}$ are input into the encoder:
\begin{equation}\label{eq:embeddings}
    z_i= 
\begin{cases}
    V^{\text{in}}_i\mathbf{E}+ \mathbf{E}^{\text{pos}}_i,& \text{if } m_i=1\\
    \textbf{E}_{\text{mask}} + \mathbf{E}^{\text{pos}}_i,              & \text{otherwise}.
\end{cases}
\end{equation}
Our encoder transforms the final embedding $\mathbf{z}$ to the full set of 3D vertex predictions $\hat{V}$. An additional layer is used to upsample to the original SMPL mesh for evaluation purposes.

\begin{figure*}
\centering
\includegraphics[width=0.75\textwidth]{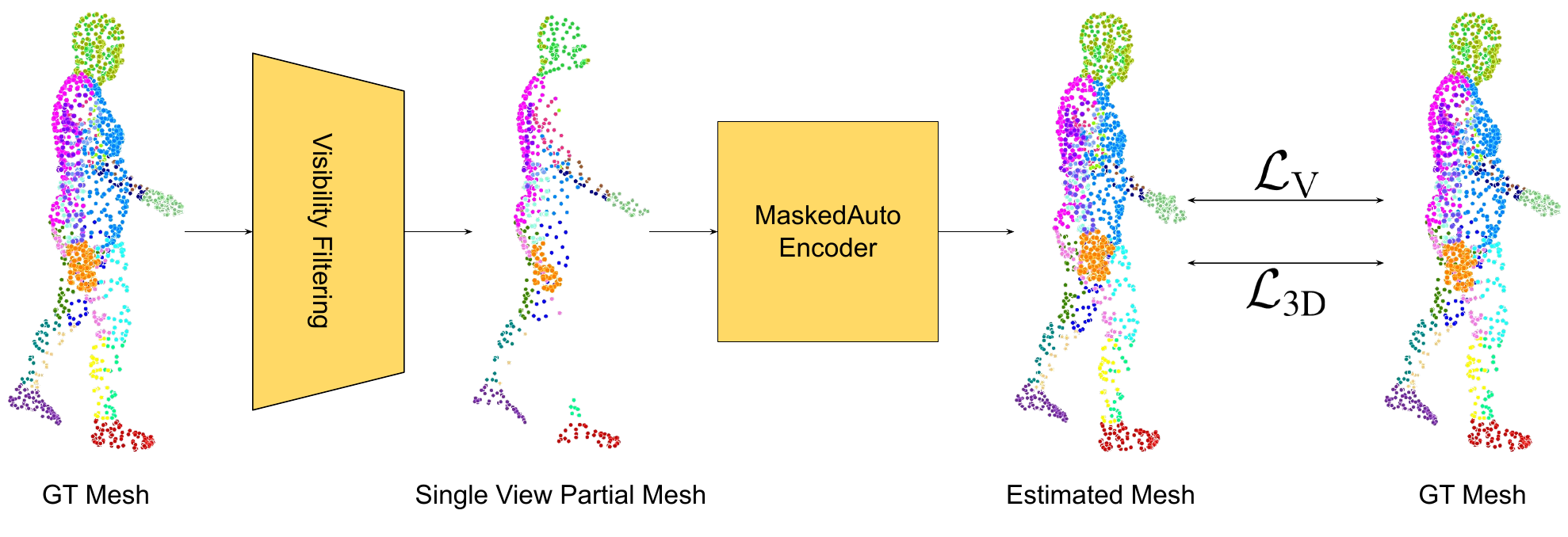}
\caption{Training pipeline of our method, M$^3$. A ground-truth body mesh, obtained from a MoCap dataset, is first downsampled and projected to a fixed virtual camera to yield the single-view partial mesh. This process simulates the ``D" channel of an RGBD camera. The resulting partial mesh is then fed to the Masked Auto Encoder (MAE) which completes the mesh. The MAE is trained using vertex loss $\mathcal{L}_\mathrm{V}$, and 3D loss $\mathcal{L}_\mathrm{3D}$.
}
\label{fig:method}
\end{figure*}

We train our encoder and learnable embedding layers with the overall loss term
\begin{equation}
    \mathcal{L} = \lambda_\text{V}\mathcal{L}_\text{V} + \lambda_\text{3D}\mathcal{L}_\text{3D}
\end{equation}
where 
\begin{equation}
    \mathcal{L}_\text{V} = \frac{1}{N}\sum_{i=1}^N||\hat{V}_i - V_i||_1,
\end{equation}
\begin{equation}
    \mathcal{L}_\text{3D} = ||\mathcal{J}\hat{V}-J_\text{3D}||_\text{F}^2
\end{equation}
where $V$ and $J_\text{3D}$ are groundtruth vertices and 3D joints, respectively, and $\mathcal{J}$ is pre-trained joint regressor matrix. Overall training pipeline of our method is given in Fig. \ref{fig:method}.

\subsection{Training Data Generation}\label{subsection:trainingdata}
The main advantage of our two-stage architecture is to alleviate the requirement for paired RGBD-mesh  data. In principle, the UV estimator $\Phi$ and our masked autoencoder $\Theta$ can be trained separately. Since we utilize DensePose for UV estimation, the only trainable module in our pipeline is the transformer-based masked autoencoder we define in Section \ref{subsection:mmm}. To this end, we utilize human meshes fitted to MoCap data to generate input pairs of vertices $V^\text{in}$, visibility mask $m$. For a target mesh $V$, we simply add noise to it to obtain the input vertices $V^\text{in}$. The visibility mask $m$ is calculated by ray casting from a synthetic viewpoint. Note that the 3D locations of non-visible vertices are replaced with mask tokens after embedding projection (Equation \ref{eq:embeddings}). 

AMASS provides us with perfect data for visible surfaces; however, the same is not true on RGB images. We bridge this domain gap partially by randomly setting a percentage of visible vertices (i.e. $m_i=1$) as non-visible. In addition, we fine-tune $\Theta$ on real data with $V^\text{in}$ generated by the UV matching step outlined in Section \ref{subsection:partialmesh}. 

%% file: sec/4_implementation_details.tex
\section{Implementation details}
\label{sec:imp_details}
We use PyTorch \cite{pytorch} to implement models  and utilize PyKeops \cite{JMLR:v22:20-275} for nearest neighbor search. We normalize input vertices to zero mean and unit variance in 3D with respect to visible ones and denormalize the encoder output. We project the normalized 3D vertex coordinates to a hidden dimension of size $20$. We employ 6 transformer blocks followed by a layer normalization and a linear layer.

During training, we additionally mask $60\%$ of vertices to align the real and simulated data. We also add Gaussian noise with zero mean and $0.0005$ variance to the input vertices. The learning rate follows a cosine decay schedule with the first $15\%$ of steps dedicated to warmup. The initial learning rate is 1e-3, and weight decay is set to 1e-4. A batch size of 32 is used during training. For fine-tuning, we disable the warmup, and the learning rate and weight decay were set to 1e-5 and 1e-6, respectively. A single training run takes 60 hours on a single NVIDIA A100 GPU. 

%% file: sec/5_experiments.tex
\section{Experiments}
\label{sec:experiments}
In this section, we present the results of our experiments, providing a thorough evaluation of our approach. To assess the accuracy and effectiveness of our method, we utilized Per Vertex Error (PVE) as the error metric. Furthermore, we conduct ablation experiments to analyze the individual contributions of different components within our methodology. We utilize two synthetic datasets, SURREAL \cite{varol17_surreal} and CAPE \cite{CAPE:CVPR:20}, offering point cloud and ground-truth SMPL pairs. 
To validate our findings in real-world scenarios, we tested our method using the BEHAVE dataset, which includes paired RGBD-SMPL data.

\subsection{Datasets}
\paragraph{Training Data.}
We employ the AMASS dataset \cite{AMASS:2019} as our training set. AMASS is a large-scale MoCap dataset with thousands of SMPL instances, providing a diverse range of motion samples as human pose and shape. It helps us to generate 
3D partial mesh and SMPL pairs as described in \cref{subsection:trainingdata}. 
We use BEHAVE's train set \cite{bhatnagar22behave}, which offers real-world RGBD-SMPL pairs, to fine-tune our model in order to adapt it to real-world scenarios.

\paragraph{Evaluation Data.}
To test our model, we utilize SURREAL \cite{varol17_surreal}, CAPE \cite{CAPE:CVPR:20}, and BEHAVE \cite{bhatnagar22behave} datasets. 
SURREAL offers RGB images, depth maps, and 3D ground-truth body meshes. We are able to generate partial mesh and 3D body mesh pairs to evaluate our model. CAPE includes clothed SMPL models, and we 
generate partial mesh and 3D body mesh pairs for evaluation.
We use BEHAVE's test set \cite{bhatnagar22behave} to test our model performance on real-world scenarios. We generate partial meshes from RGBD frames and evaluate our model with these RGBD-generated partial meshes and 3D human body mesh pairs.

\begin{table}[]
\centering
\begin{tabular}{|l|l|c|}
\hline
\textbf{Method}                                                        & \textbf{Input} & \textbf{PVE} \\ \hline \hline
DynaBOA$^\dagger$ \cite{guan2022out} \hfill \textcolor{gray}{(PAMI21)} & $RGB$          & 70.7             \\ \hline
PC-HMR \cite{luan2021pc} \hfill \textcolor{gray}{(AAAI21)}             & $RGB$          & 59.8             \\ \hline
Ma et al. \cite{ma20233d} \hfill \textcolor{gray}{(CVPR23)}            & $RGB$          & 44.7             \\ \hline \hline
Jang et al. \cite{jang2023dynamic} \hfill\textcolor{gray}{(ICCV23)}    & $D^*$          & 13.2             \\ \hline \hline
3DCODED \cite{groueix20183d} \hfill\textcolor{gray}{(ECCV18)}          & $D$            & 41.8             \\ \hline
Wang et al. \cite{wang2020sequential} \hfill\textcolor{gray}{(CVPR20)} & $D$            & 24.3             \\ \hline
SyNoRiM \cite{huang2022multiway}  \hfill\textcolor{gray}{(PAMI22)}     & $D$            & 21.9             \\ \hline
VoteHMR \cite{liu2021votehmr} ~~~\hfill\textcolor{gray}{(ACMMM21)}     & $D$            & 20.2             \\ \hline
PointHPS \cite{cai2023pointhps}  \hfill\textcolor{gray}{(2023)}        & $D$            & 19.6             \\ \hline \hline
M$^3$ (Ours)                                                           & $RGB+D$        & \textbf{16.8}    \\ \hline
\end{tabular}
\caption{Comparison with other methods on the SURREAL dataset.  The values are measured using the PVE metric in mm. $RGB$ represents single view $RGB$, $D$ represents single-view Depth, $PC$ represents complete human Point Cloud, and (*) denotes method that take temporal data as input. $(\dagger)$ denotes method training on the test split with 2D supervisions.}
\label{table:surreal_main}
\end{table}

\begin{table}[]
\centering
\begin{tabular}{|l|l|c|}
\hline
\textbf{Method}                                                    & \textbf{Input} & \textbf{PVE} \\ \hline \hline
IP-NET \cite{bhatnagar2020ipnet} \hfill\textcolor{gray}{(ECCV20)}  & $PC$             & 28.2          \\ \hline
PTF \cite{wang2021locally}  \hfill\textcolor{gray}{(CVPR21)}       & $PC$             & 23.1          \\ \hline \hline
IP-NET \cite{bhatnagar2020ipnet} \hfill\textcolor{gray}{(ECCV20)}  & $D$              & 60.2          \\ \hline
PTF \cite{wang2021locally}  \hfill\textcolor{gray}{(CVPR21)}       & $D$              & 52.3          \\ \hline \hline
M$^3$ (Ours)                                                       & $RGB+D$          & \textbf{22.0} \\ \hline
\end{tabular}
\caption{Comparison with other methods on the CAPE dataset.  The values are measured using the PVE metric in mm. $D$ represents single-view Depth, $PC$ represents complete human Point Cloud as input.}
\label{table:cape_main}
\end{table}

\subsection{Quantitative Results}

We train our model on the AMASS dataset in our experiments and assessed its performance on multiple datasets, including SURREAL, CAPE, and BEHAVE. SURREAL and CAPE datasets provide synthetic data, while BEHAVE comprises real-world samples. We conduct a comparative analysis using several methods that use single-view depth, temporal single-view depth, and complete human point cloud as input.

In \cref{table:surreal_main}, we compare our method with other methods that use single-view depth as input on the SURREAL dataset in terms of PVE. Our method outperforms other single-frame methods, showcasing the effectiveness of our method and the benefits of utilizing partial meshes (formed through our UV matching using the RGB channels). Our method outperforms PointHPS \cite{cai2023pointhps}, the closest competitor that uses monocular point cloud, by $2.8$mm. However, Jang et al. \cite{jang2023dynamic} leverages temporal information and obtains better results on the PVE metric compared to ours. As the SURREAL dataset is synthetic, it lacks noise in depth frames, which benefits the performance of all methods. To highlight the robustness of our model, discussed in \cref{sub:noise}, we introduced noise to the inputs. Additionally, our method also outperforms RGB-based methods, as shown in the table.

We also compare our method with IP-NET \cite{bhatnagar2020ipnet} and PTF \cite{wang2021locally}—methods capable of handling both single-view depth and complete human point cloud as input within the CAPE dataset in terms of PVE metric, as depicted in \cref{table:cape_main}.
Our method significantly outperforms others in single-view depth settings and even demonstrates superior performance when they utilize complete human point clouds. Specifically, our method improves PVE by $30.3$mm compared to PTF \cite{wang2021locally} in the single-view depth setting. Moreover, despite using only single-view depth, our method achieves $1.1$mm improvement in PVE compared to PTF using complete human point clouds as input.
In addition, both IP-NET \cite{bhatnagar2020ipnet} and PTF \cite{wang2021locally} utilize optimization loops to obtain final results, which significantly extends the time required for the process. In contrast, our model works in a single pass without using loops, making it much more efficient.

We evaluate our model in real-world scenarios using the BEHAVE \cite{bhatnagar22behave} dataset.  Despite employing single-view RGBD compared to others using complete human point clouds or multiview RGBD frames, our method demonstrates competitive performance as shown in \cref{table:BEHAVE_comparison}. Our approach showcases significantly superior performance compared to LoopReg \cite{bhatnagar2020loopreg}, even when provided with a complete point cloud. Our method shows relatively lower performance than IP-Net, which utilizes a complete human point cloud. We compare our model with a naive baseline, which directly fits the partial mesh to the SMPL body model using $\mathcal{L}_\text{v2v}$ loss. 
On the other hand, our method is single pass approach while LoopReg\cite{bhatnagar2020loopreg}, BEHAVE \cite{bhatnagar22behave}, and IPNet \cite{bhatnagar2020ipnet} uses an optimization loop at test time. 
We compared our method with the most recent RGB-based method TokenHMR \cite{dwivedi_cvpr2024_tokenhmr}. Our method outperforms TokenHMR by $18.4$ mm in PVE metric. This improvement demonstrates the advantage of incorporating depth information along with RGB data. We also evaluated TokenHMR using the Mean Per Joint Position Error (MPJPE) metric, resulting in $71.2$ mm, whereas our method achieved $57.9$ mm. Our method improves MPJPE by $13.3$ mm compared to TokenHMR.

Our model exhibits a performance decline on the BEHAVE dataset in comparison to SURREAL and CAPE. The BEHAVE dataset involves subjects with loose clothing, impacting DensePose predictions by causing predictions to extend beyond the actual body boundaries. As our model relies on DensePose predictions, its performance decreases when applied to subjects with loose fitting.

\begin{table}[h]
\centering
\begin{tabular}{|l|l|c|}
\hline
\textbf{Method}  & \textbf{Input} & \multicolumn{1}{l|}{\textbf{PVE}} \\ \hline \hline
Baseline (Opt. w/ $\mathcal{L}_\text{v2v}$) & $RGB$+$D$ & 145.9             \\ \hline
PHOSA \cite{zhang2020phosa}  \hfill\textcolor{gray}{(ECCV20)}       & $RGB$               & 137.3             \\ \hline
LoopReg \cite{bhatnagar2020loopreg} \hfill\textcolor{gray}{(NIPS20)}        & Multiview $D$       & 91.2                                 \\ \hline
TokenHMR \cite{dwivedi_cvpr2024_tokenhmr}  \hfill\textcolor{gray}{(CVPR24)}        & $RGB$        & 89.3                        \\ \hline
IP-NET\cite{bhatnagar2020ipnet} \hfill\textcolor{gray}{(ECCV20)}   & $PC$             & 66.1                                 \\ \hline
BEHAVE \cite{bhatnagar22behave}  \hfill\textcolor{gray}{(CVPR22)}        & Multiview 
 $D$        & \textbf{49.9}                        \\ \hline \hline
M$^3$(Ours) & $RGB$+$D$  & 70.9                                \\ \hline
\end{tabular}
\caption{Comparison with other methods on BEHAVE dataset. $D$:  single-view depth, $PC$:  complete human point cloud. 
}
\label{table:BEHAVE_comparison}
\end{table}

\begin{table}[h]
\centering
\begin{tabular}{ |l|c|c|c| }
\hline 
\textbf{Method}   & \textbf{PVE}    & \textbf{MPJPE}  \\ \hline \hline
M$^3$ & 93.1 & 76.2  \\ \hline
M$^3$ fine-tuned & 70.9   & 57.9 \\ \hline
\end{tabular}
\caption{Effect of fine-tuning  on the BEHAVE dataset.}
\label{table:baseline_finetuning}

\end{table}

\subsubsection{Ablation of Fine-tuning}

Given that the BEHAVE dataset is comparatively smaller in scale compared to MoCap-generated ground-truth SMPL datasets, we chose to leverage the AMASS dataset for training our network. Our model achieves respectable performance using only the AMASS dataset. However, due to the nature of the AMASS dataset not containing any RGB data, 
where we generate partial point clouds 
by projecting meshes onto a fixed camera, we extend our approach to include the BEHAVE train set for training. This adaptation results in a significant improvement in performance on the BEHAVE test set. The impact of fine-tuning on the BEHAVE dataset is shown in \cref{table:baseline_finetuning}. The fine-tuning process improved our PVE result from $93.1$mm to $70.9$mm and MPJPE results from $76.2$mm to $57.9$mm.

\subsubsection{Robustness to Varied Noise Levels} \label{sub:noise}

Depth sensor noise is a common challenge. To evaluate the robustness of our method to noisy measurements, we introduced varying levels of noise to the inputs in the SURREAL dataset to simulate different real-world noise scenarios and compared our performance to VoteHMR \cite{liu2021votehmr} (Fig. 4 in supplementary material). Each point in a test sample was offset by random 3D noise according to a given standard deviation, similar to the approach in VoteHMR. Our method demonstrated robustness to different noise levels, showing only marginal increases in the PVE metric compared to VoteHMR. We also tested our model's robustness on the CAPE dataset by applying various types of noise to the samples. As shown in \cref{table:noise}, our model exhibited consistent performance across varying noise levels without significant degradation, demonstrating the robustness of the masked autoencoder.



\begin{table}[]
\centering
\begin{tabular}{|c|c|c|c|c|}
\hline
\textbf{Std of Noise (mm)} & 0    & 10   & 30   & 50   \\ \hline \hline
\textbf{PVE}   & 22.0 & 22.3 & 25.0 & 30.2 \\ \hline
\end{tabular}
\caption{Effect of noise on the CAPE dataset on Per Vertex Error in millimeters.}
\label{table:noise}
\end{table}

\begin{figure*}
\centering
\includegraphics[width=1.2\columnwidth]{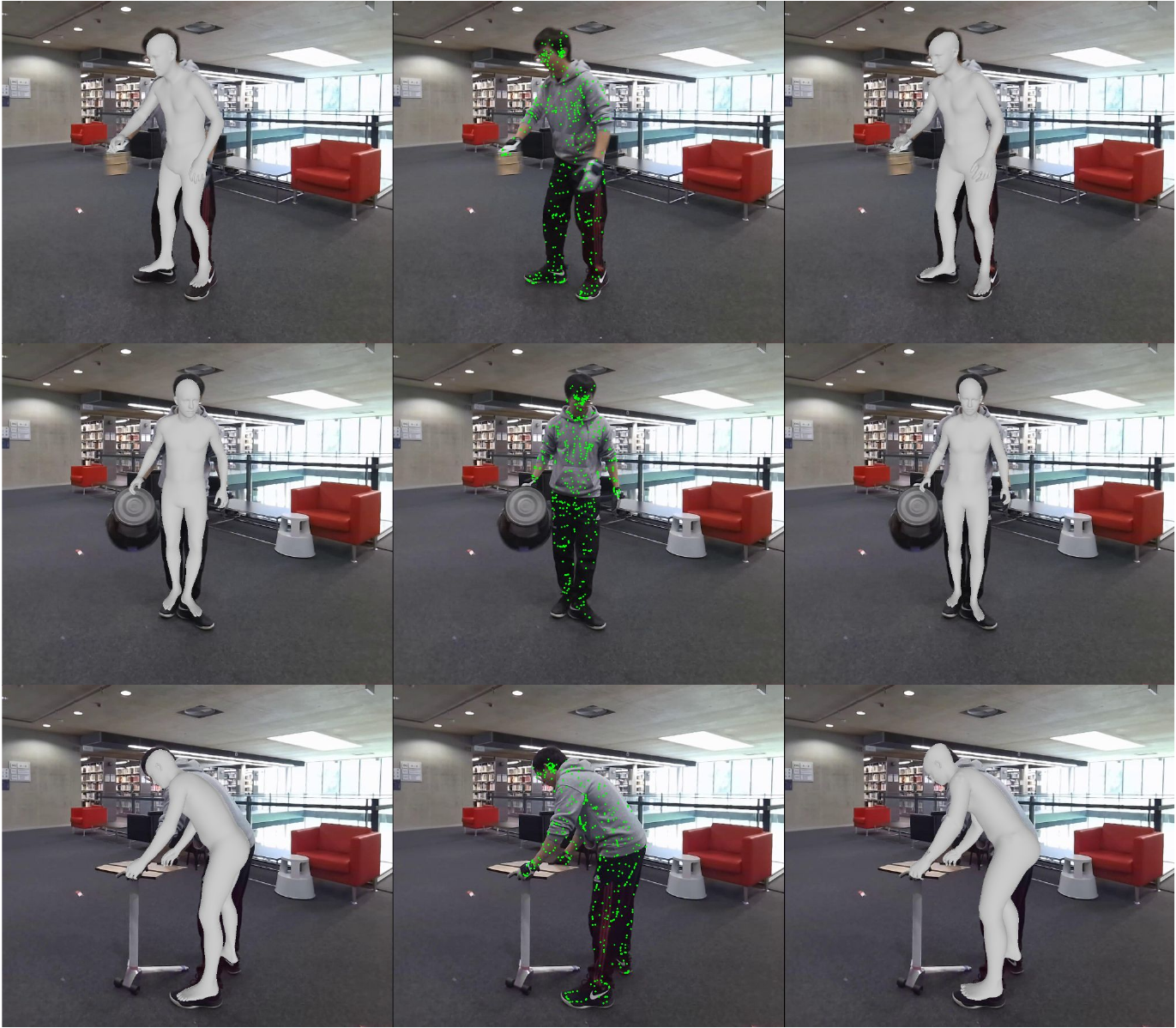}
\caption{Visual results on BEHAVE test set. From left to right: ground truth, model inputs from RGBD, and model output. }
\label{fig:qualitative_behave}
\end{figure*}

\begin{figure}
\centering
\includegraphics[width=\columnwidth]{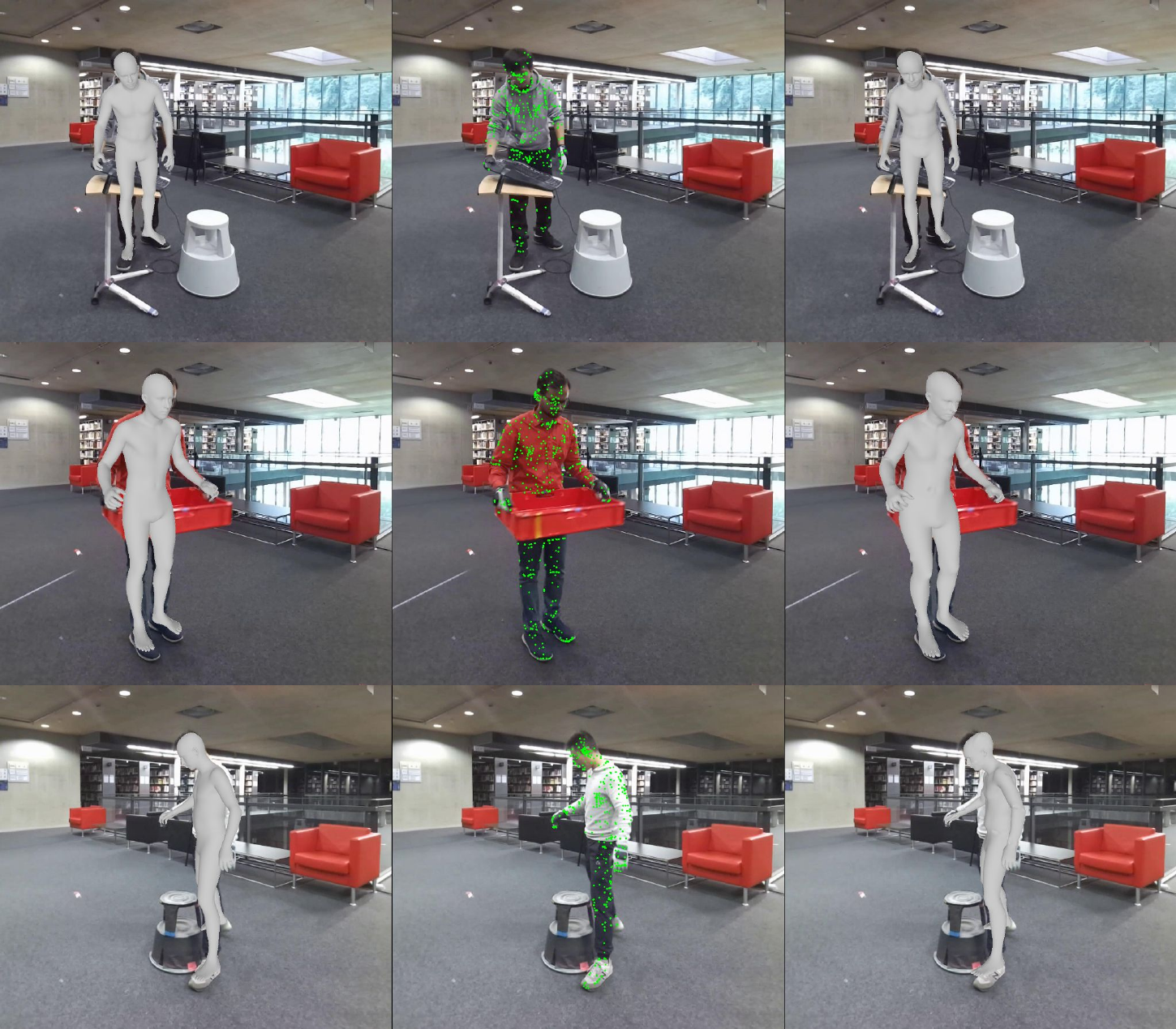}
\caption{Occlusion examples from  BEHAVE test set. The first two rows show examples where the subject is occluded by objects. The third row illustrates a case of self-occlusion:  the subject's right leg is occluded by the left leg but it is correctly recovered by our model. From left to right: ground truth, model inputs from RGBD, and model output.}
\label{fig:occlusion}
\end{figure}

\begin{figure}
\centering
\includegraphics[width=\columnwidth]{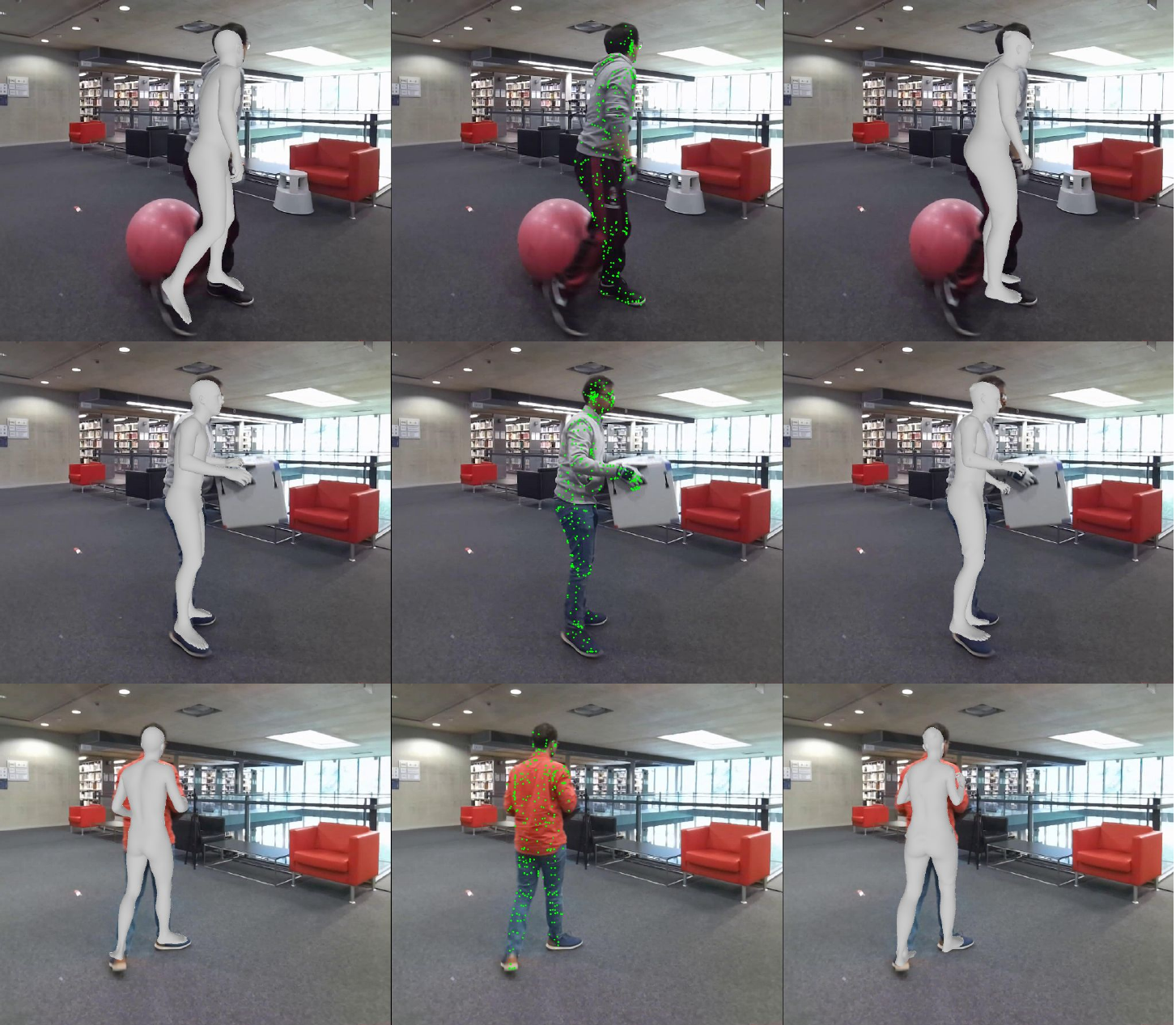}
\caption{Failure cases on BEHAVE test set. From left to right: ground truth, model inputs from RGBD, and model output.}
\label{fig:qualitative_render_limitations}
\end{figure}

\subsection{Qualitative Results}

Selected samples from our qualitative results are shown in \cref{fig:qualitative_behave}, demonstrating our method's effectiveness in real-world scenarios.
The images showcased effectively demonstrate our model's ability to estimate and complete unseen parts of the human body. The samples are taken from the BEHAVE test set. Our method successfully reconstructs the 3D human mesh when a significant portion of the body is visible. Moreover, in occlusion cases where only a small portion of a body part is visible, our model effectively completes the occluded section of the corresponding body part. As shown in \cref{fig:qualitative_behave}, row 3 is an example where our model accurately captures the subject's right arm from just a few visible points.

However, it's important to acknowledge the limitations of our model, which we illustrate in \cref{fig:qualitative_render_limitations}. In the first row, the subject's right foot is in motion, resulting in motion blur at the right lower leg and foot. In such cases, DensePose struggles to capture these body parts from the RGB data, preventing us from accurately reconstructing depth points for those areas. This leads to errors in the reconstruction of the subject's right leg. Additionally, heavy occlusions at the body's extremities can result in errors in the corresponding parts, as seen in rows 2 and 3; both have errors on the subject's left arm.

%% file: sec/6_conclusion.tex
\section{Conclusions}
\label{sec:conclusion}
In this study, we address the challenging task of accurate 3D human mesh estimation from single-view RGBD data, focusing on the critical problem of completing a 3D partial human body mesh. 

We propose Masked Mesh Modeling ($M^3$), inspired by masked image modeling, to reconstruct missing points and recover full 3D meshes.

Due to the scarcity of paired RGBD-3D human body mesh data, we utilize Motion Capture (MoCap) data to create partial mesh and 3D human body mesh pairs. Following training with MoCap data, we fine-tune our model using real-world RGBD-SMPL data to enhance its applicability to real-world scenarios.

We evaluated the performance of our model on both real-world and synthetic datasets. On synthetic datasets, our approach outperforms existing single-view and multi-view methods, yielding the most favorable results. We demonstrated the robustness of our model to noise by introducing various noise levels to synthetic inputs. Furthermore, our method exhibits competitive performance on real-world datasets, underscoring its efficacy in practical scenarios. As RGBD sensors become increasingly accessible, our approach becomes applicable to real-world problems at a reduced cost.

\subsection{Future Work and Limitations}


The current method uses a single-view RGBD camera and generates 3D body points with the help of depth information from the camera. Our method generates an SMPL body from a partial set of human body points. Partial points can be obtained from multiview calibrated RGB cameras by triangulating DensePose estimations from multiple views. This process will take more time than our method, but there may be an increase in accuracy since partial points will come from different viewpoints. RGBD camera detects points only from one view.

We plan to extend our method to use RGB inputs alone by estimating depth from RGB data. This will allow mesh estimation without relying on RGBD sensors, broadening the applicability of our method.

\section*{Acknowledgements}
We gratefully acknowledge the computational resources kindly provided by TÜBİTAK ULAKBİM High Performance and Grid Computing Center (TRUBA). Dr. Akbas gratefully acknowledges the support of TUBITAK 2219.

%% file: sec/X_supp.tex
\clearpage
\setcounter{page}{1}
\setcounter{figure}{0}
\setcounter{section}{0}
\maketitlesupplementary

\section{Extra Qualitative Results}
\label{sec:supp_qualitative}

\begin{figure}
    \centering
    \includegraphics[width=0.9\columnwidth]{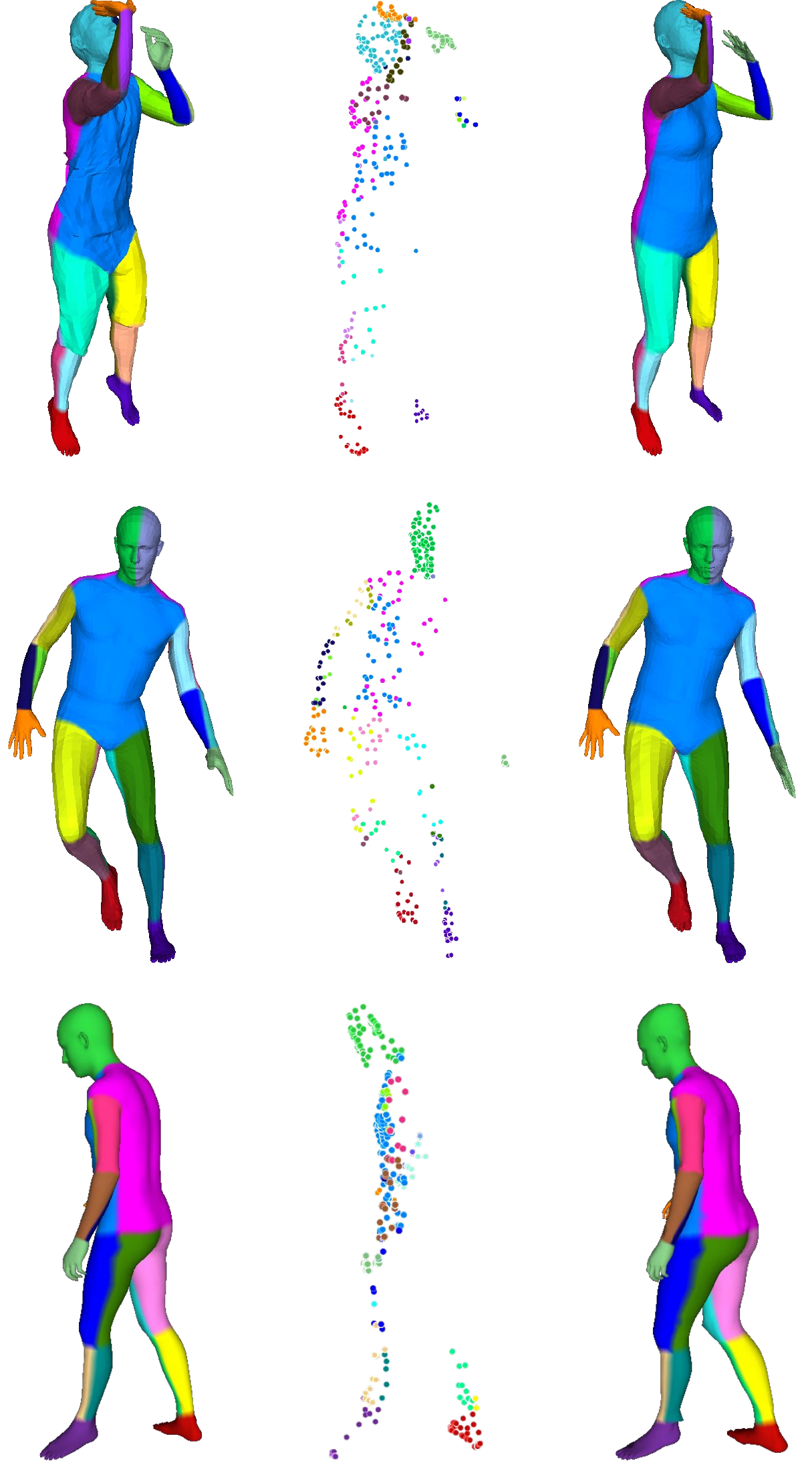}
    \caption{Estimations from CAPE \cite{CAPE:CVPR:20} (First Row), SURREAL\cite{varol17_surreal} (Second Row), and BEHAVE\cite{bhatnagar22behave} (Third Row). The first column is ground truth, the second column is model input, and the third column is model output.}
    \label{fig:accurate}
\end{figure}

We have added some more qualitative results. \cref{fig:accurate} displays successful outcomes of the M$^3$ model. Our model can reconstruct partially visible body parts. In the first two rows, the model accurately completes the left arm of the subjects. In these examples, the upper arm orientation is estimated from lower arm information, and the model successfully completes the upper arms. The third row illustrates the successful completion of the right leg, similar to the first two rows.

\begin{figure}
    \centering
    \includegraphics[width=0.9\columnwidth]{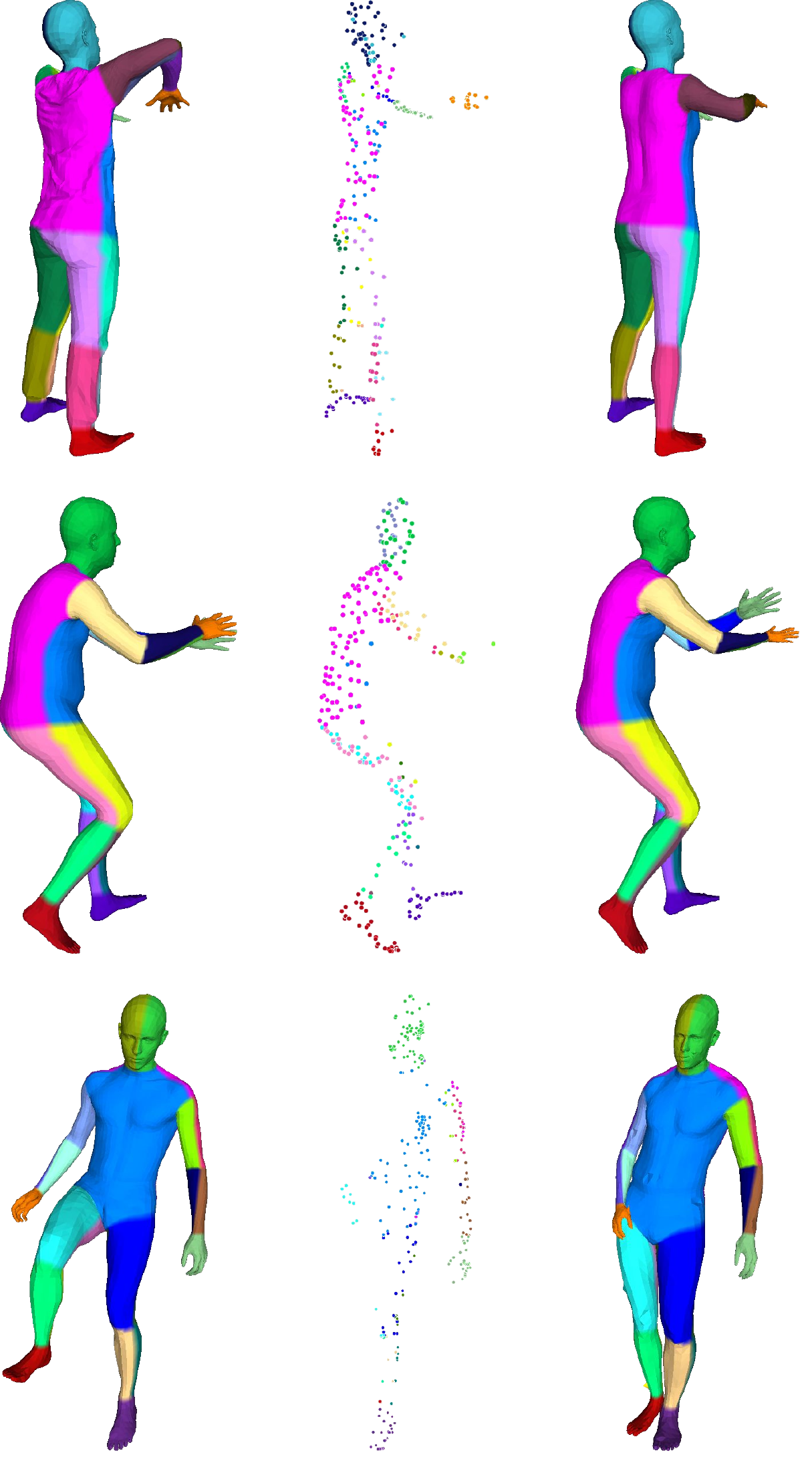}
    \caption{Estimations with minor errors from CAPE (First Row), SURREAL (Second Row), and BEHAVE (Third Row). The first column is ground truth, the second column is model input, and the third column is model output.}
    \label{fig:small_error}
\end{figure}

\cref{fig:small_error} shows instances of minor errors. When the model lacks information about a specific body part, it may reconstruct a random orientation that aligns with visible points. In the first row, the right elbow is estimated solely based on the right hand location, and the elbow can be located in more than one place for the given input. The second row presents a similar problem. The model makes errors when estimating hand locations due to insufficient information in the input. In the third row, where the model doesn't observe any points from the right side of the body, estimations for the right body parts are inaccurate.

\begin{figure}
    \centering
    \includegraphics[width=0.9\columnwidth]{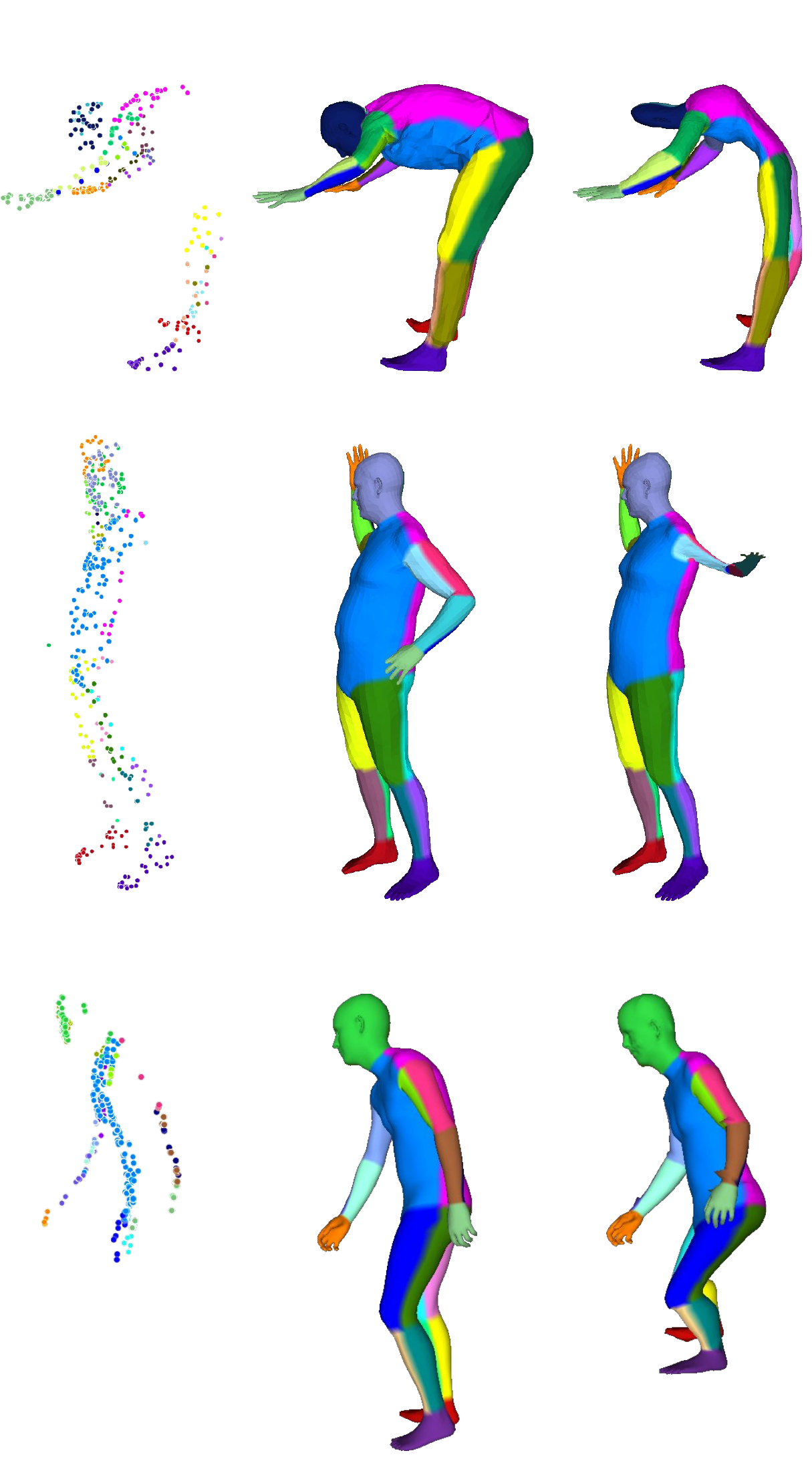}
    \caption{Estimations with significant errors from CAPE (First Row), SURREAL (Second Row), and BEHAVE (Third Row). The first column is ground truth, the second column is model input, and the third column is model output.}
    \label{fig:erroneous}
\end{figure}

The model makes significant errors on some challenging examples, as in the \cref{fig:erroneous}. In the first case, where the middle of the body lacks any points, the model struggles to reconstruct the hip region. For body parts with only noisy sparse points, as shown in the second row, the model produces distorted shapes for those body parts. The final row shows an issue with non-visible body part reconstruction. The partial body lacks leg points, causing the model to generate shorter legs to align with dense points near the hips.

We also generated several video outputs from the BEHAVE test set. You can access the videos using the following link: \href{https://tinyurl.com/maskedmeshmodelling}{tinyurl.com/maskedmeshmodelling}. Each frame of the video contains 3 columns. The first column is the ground truth, the second column is the model input, and the third column is the output.

\section{Robustness to Noise}
\label{sec:supp_noise}
\begin{figure}
\centering
\includegraphics[width=\columnwidth]{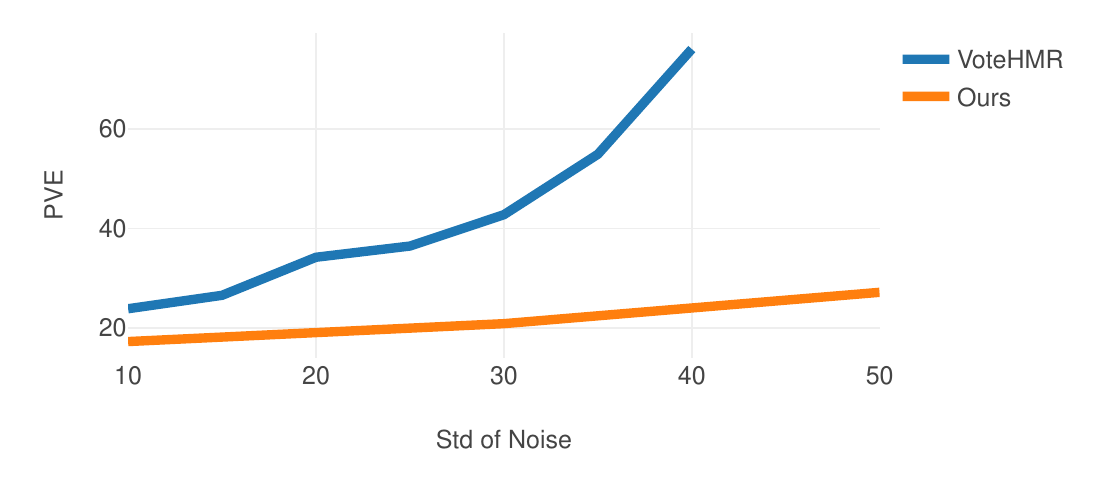}
\caption{Comparison between our method M$^3$ and VoteHMR \cite{liu2021votehmr} regarding the effect of noise on Per Vertex Error (PVE) in mm.}
\label{fig:noise}
\end{figure}

We tested noise robustness by applying varying noise levels to the input, similar to VoteHMR \cite{liu2021votehmr}. \cref{fig:noise} compares our method with VoteHMR. Our method demonstrates more consistent results compared to VoteHMR as the noise level increases.

